\documentclass[conference]{IEEEtran}
\IEEEoverridecommandlockouts
\usepackage{cite}
\usepackage{amsmath,amssymb,amsfonts}
\usepackage{algorithmic}
\usepackage{graphicx}
\usepackage{textcomp}
\usepackage{xcolor}
\usepackage{hyperref} 

\def\BibTeX{{\rm B\kern-.05em{\sc i\kern-.025em b}\kern-.08em
    T\kern-.1667em\lower.7ex\hbox{E}\kern-.125emX}}
\begin{document}

\title{Forearm Ultrasound based Gesture Recognition on Edge\\
}

\author{\IEEEauthorblockN{Keshav Bimbraw}
\IEEEauthorblockA{\textit{Robotics Engineering} \\
\textit{Worcester Polytechnic Institute}\\
Worcester, USA \\
kbimbraw@wpi.edu}
\and
\IEEEauthorblockN{Haichong K. Zhang}
\IEEEauthorblockA{\textit{Robotics Engineering and Biomedical Engineering} \\
\textit{Worcester Polytechnic Institute}\\
Worcester, USA \\
hzhang10@wpi.edu}
\and
\IEEEauthorblockN{Bashima Islam}
\IEEEauthorblockA{\textit{Electrical and Computer Engineering} \\
\textit{Worcester Polytechnic Institute}\\
Worcester, USA \\
bislam@wpi.edu}
}

\maketitle

\begin{abstract}
Ultrasound imaging of the forearm has demonstrated significant potential for accurate hand gesture classification. Despite this progress, there has been limited focus on developing a stand-alone end- to-end gesture recognition system which makes it mobile, real-time and more user friendly. To bridge this gap, this paper explores the deployment of deep neural networks for forearm ultrasound-based hand gesture recognition on edge devices. Utilizing quantization techniques, we achieve substantial reductions in model size while maintaining high accuracy and low latency. Our best model, with Float16 quantization, achieves a test accuracy of 92\% and an inference time of 0.31 seconds on a Raspberry Pi. These results demonstrate the feasibility of efficient, real-time gesture recognition on resource-limited edge devices, paving the way for wearable ultrasound-based systems.
\end{abstract}


\section{Introduction}
For effective human-machine interfacing, it is necessary to develop sensing techniques that do not interfere with the natural movements of hands. Several techniques such as vision-based methods, wearable sensor gloves, and depth sensor-based approaches have been proposed to estimate hand movements. However, these are sensitive to light conditions, shadows, occlusion, and usually prevent free movement of hands \cite{b1}. Biosignal based modalities can be used to understand hand movements, and several such techniques have been used in the literature, including but not limited to electromyography (sEMG), force myography (FMG), ultrasound, and photoplethysmography (PPG) \cite{b2}. There are several pros and cons to every modality, however, ultrasound, specifically forearm ultrasound is a unique topic of interest because it provides a high-dimensional visualization of the forearm musculature. Visualizing the physiological changes in the forearm can be used to infer what is happening in the hand. Recent work has shown that it can be used to effectively measure hand gestures, finger angles, and finger forces \cite{b3, b4, b5, b6}. 

Most of the research in forearm ultrasound-based hand gesture estimation is primarily based on offline analysis of ultrasound images. This is usually done using signal processing and deep learning techniques on powerful desktop systems, usually with GPUs. Brighness-Mode (B-Mode) data refers to two-dimensional (2-D) ultrasound images that visualize internal tissue structures by displaying the intensity of reflected sound waves as varying levels of brightness. While the hand gesture classification performance based on 2-D B-Mode data obtained using ultrasound probes is promising as shown in \cite{b7}, these approaches are impractical for real-time applications due to their reliance on bulky, high-power systems. This study aims to bridge this gap by demonstrating gesture recognition on resource-constrained edge devices. Achieving this presents significant challenges, including the need to efficiently process high-dimensional ultrasound data with limited computational resources and ensuring low latency for practical usability.

The importance of this technology lies in its potential to enable real-time hand gesture recognition in an online setting, enhancing human-machine interfaces and making the system more wearable and practical. For ultrasound based gesture recognition, it is crucial for several reasons: it ensures user privacy by processing data locally without transmitting it to external servers, thereby reducing the risk of data breach. Additionally, edge devices are typically more power-efficient, which is essential for wearable applications that require long battery life. Advances in on-device deep learning also facilitate user personalization and continuous model improvement, allowing for adaptive systems that can learn and evolve over time \cite{b8}. Furthermore, deploying models on edge devices supports the implementation of smaller and faster models which can be beneficial in terms of latency and performance.

\begin{figure*}
  \centering 
\includegraphics[width=440pt]{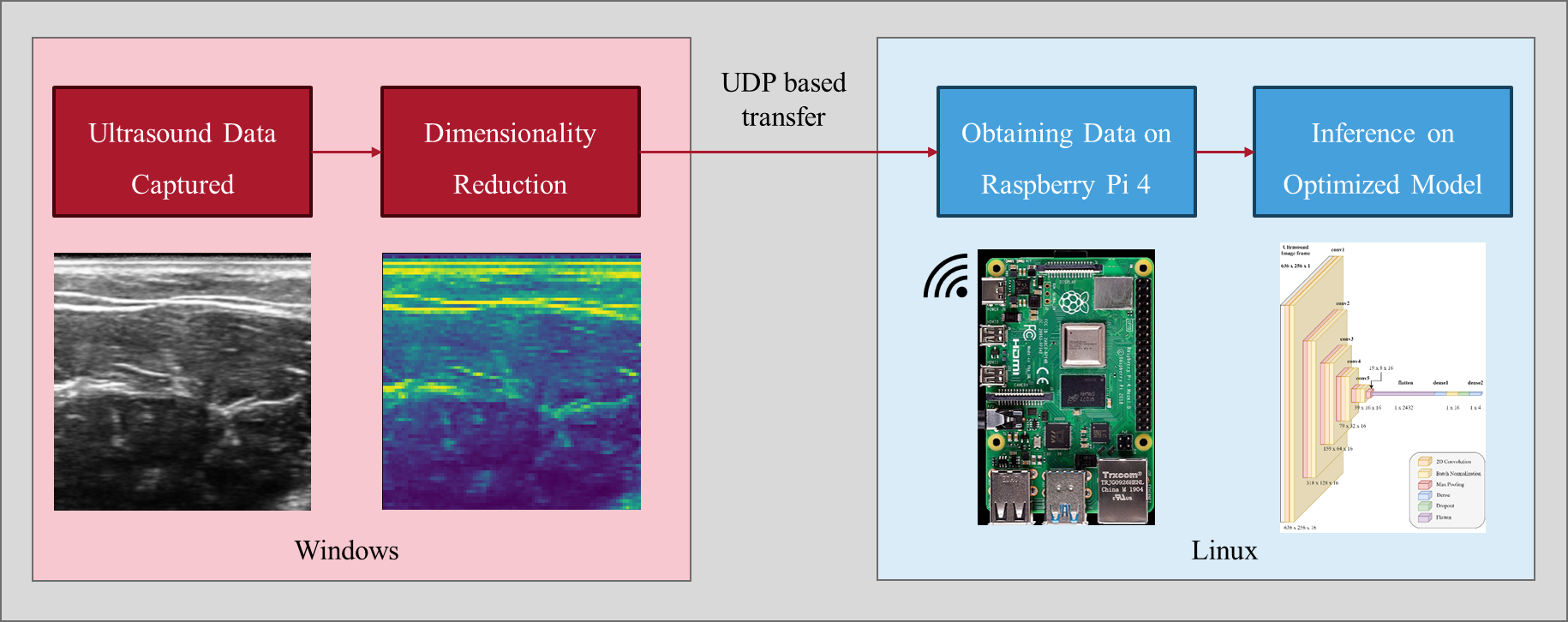}
  \caption{Schematic for ultrasound based gesture estimation deployed on a Raspberry Pi. The ultrasound image is obtained from an ultrasound probe, and it is sent to the Raspberry Pi over UDP after downsizing. Here, a pre-trained and optimized model is used for inference.}
  \label{fig:teaser}
\end{figure*}

We addressed the challenge of deploying a deep learning model for ultrasound based hand gesture recognition on a resource-constrained edge device, specifically a Raspberry Pi. Our approach involved training a Convolutional Neural Network (CNN) model, followed by quantization to reduce its size while maintaining performance. Novel aspects include the deployment of the compressed model for real-time inference on the Raspberry Pi. Ultrasound data was downsized on a Windows system and sent to Raspberry Pi via Universal Datagram Protocol (UDP) for inference. Our findings showed that with Float16 quantization, the model achieved a test accuracy of 92\% and an inference time of 0.31 seconds, reducing the model size from 1.5 megabytes to 0.24 megabytes. Dynamic range quantization further reduced the model size to 0.12 megabytes with a slight performance trade-off. This setup is illustrated in Figure \ref{fig:teaser}.

\section{Experimental Framework and Implementation}
This section details the comprehensive framework and implementation of our on-device deep learning approach for gesture recognition using ultrasound data. We utilized a SonoQue L5 linear probe to acquire ultrasound data from a single subject, approved under IRB-23-0634. The probe was securely strapped to the subject’s forearm using a custom-designed 3D-printed wearable armband.

\subsection{Data Collection}
\begin{figure}[b]
  \centering
  \includegraphics[width=240pt]{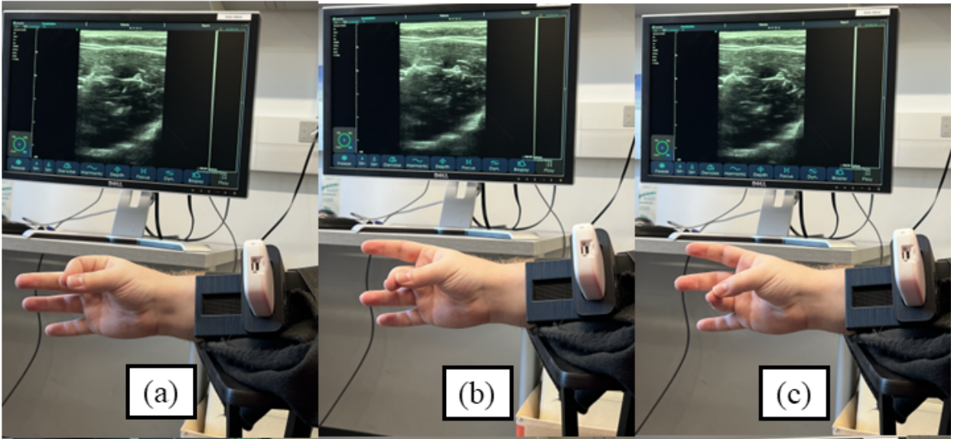}
  \caption{Hand gestures: (a) Index Pinch, (b) Middle Pinch, (c) Ring Pinch, and Open Hand (not shown).}
  \label{fig:gestures}
\end{figure}

The data comprised 4 gestures, namely, open hand gesture, and finger pinch gestures. The latter was defined to be an index-thumb pinch, middle-thumb pinch, and ring-thumb pinch. The gestures are shown in Figure \ref{fig:gestures} with the corresponding ultrasound image output on a screen. Each image was 640 x 640 pixels with total 2400 frames corresponding to 8 rounds of data acquisition (300 frames per round) with an average frame rate of 10 Hz. The data was subjected to 25\% test-train split, leading to 1800 train samples and 600 test samples.

\subsection{CNN Architecture and Training}
The network architecture is based on the CNN proposed in \cite{b4, b5}. The network uses 5 cascaded convolution layers with normalization and max-pooling, followed by flattened and dense layers. The output is 4 class probabilities for each of the gestures for a given ultrasound image. Adam optimizer and a learning rate of 1e-3 were used. The models were trained for 25 and 50 epochs for baseline performance analysis, and 20 epochs for deploying and demonstrations. All the development for training was done on a Windows 10 system, with the models trained on an NVIDIA GeForce RTX 2070 SUPER GPU using TensorFlow and Python \cite{b7}.

\begin{figure*}
  \centering
  \includegraphics[width=400pt]{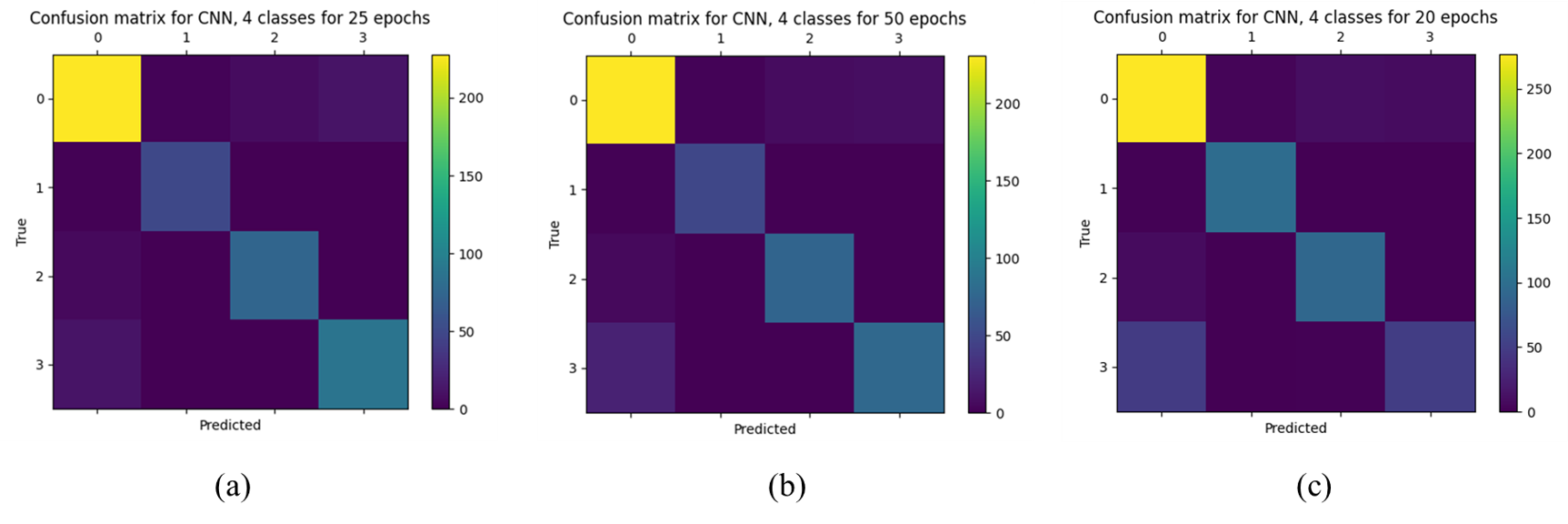}
  \caption{Confusion matrices showing the results for baseline classification performance on the test-set for (a) 25 epochs, and for (b) 50 epochs. Table showing the results for different types of quantizations with size in bytes.}
  \label{fig:confusion}
\end{figure*}

\subsection{System Description}
We utilized a Raspberry Pi 4 running Debian for our on-device deep neural network system. The trained models were converted to TensorFlow Lite (TFLite) models using TensorFlow Lite \cite{b10} to facilitate efficient on-device inference. Communication between the Raspberry Pi and the Windows system was achieved using the User Datagram Protocol (UDP). This setup allowed ultrasound images to be initially captured on the Windows system, downsampled, and then transmitted to the Raspberry Pi via UDP. The original 640 x 640 images were downsampled to 80 x 80, reducing the data size by a factor of 8. This downsampling was necessary to ensure smooth communication within the constraints of UDP's data transfer limits, avoiding buffer overflow and minimizing data loss and transfer time. In this configuration, the Raspberry Pi acted as the server, while the Windows machine functioned as the client, capturing and sending ultrasound data to the Raspberry Pi as illustrated in Figure \ref{fig:teaser}.

\subsection{Model Quantization}
To implement model compression, we employed quantization as the primary approach, exploring three distinct techniques: Float16 quantization, Dynamic Range Quantization, and UInt8 quantization \cite{b11}. Initially, the input tensors to the network were in Float32 format. During Float16 quantization, weights were converted to 16-bit floating point values during the conversion of the model from TensorFlow to TensorFlow Lite's flat buffer format. In Dynamic Range Quantization, weights were quantized post-training, while activations were dynamically quantized during inference. Notably, the model weights were not retrained to mitigate quantization-induced errors in this technique. UInt8 quantization approximates floating-point values by subtracting the zero-point value from the int8 value and multiplying the result by a scaling value. Each quantization method exhibits specific behaviors in terms of accuracy degradation and model compression.
\subsection{Evaluation}
The size of the models is described in bytes and megabytes (MB). Accuracy percentage was used as the metric to evaluate the performance of the trained model for both offline and online estimation.

\section{Results and Discussion}
In this section, we present experimental results and discussions on deploying deep learning models for ultrasound-based gesture recognition on edge devices. We analyze baseline performance, model optimization, quantization techniques, inference latency, and evaluate the deployment of the optimized model on Raspberry Pi for inference on test data.
\subsection{Baseline Performance Analysis}
Using the CNN, the baseline performance obtained on the GPU is shown in Figure \ref{fig:confusion}(a) and (b). After 25 epochs, the training data achieved an accuracy of 92.3\%, while the accuracy on the test data reached 91.5\%. Upon extending training to 50 epochs, the accuracy on the training data increased to 96\%, yet the accuracy on the test data slightly decreased to 90\%. Despite the improvement in training performance, there was a marginal decline in test performance, indicating potential overfitting to the training data, considering the independence of the train and test sets.
\subsection{Model Optimization and Deployment}
For the final analysis and demonstration, we employed a separate model trained for 20 epochs, which was subsequently optimized for UDP-based transfer. The results for this model are shown in Figure \ref{fig:confusion}(c). For this model, the train accuracy reached 93\% and the test accuracy was 92\%.
\subsection{Model Size Reduction and Quantization}
It was observed that upon employing the tflite converter, the model size decreased significantly from $\sim$1.5 MB to $\sim$0.46 MB. Subsequently, with the implementation of Float16 quantization, the size was further reduced to approximately 0.24 MB. Dynamic range quantization and UInt8 quantization resulted in an additional halving of the model size. However, the training and testing accuracy significantly dropped with UInt8 quantization. Conversely, for the other techniques, the accuracy remained high at 96\% for training and 92\% for testing. Notably, dynamic range quantization led to a marginal decrease in accuracy by 1%.
\subsection{Inference Latency Analysis}
In terms of inference latency, the highest latency was obtained for dynamic range quantization and the lowest for the UInt8 quantization. Although UInt8 quantization exhibited the least latency for inference on the test set, its poor performance necessitated consideration of the next best option, which was Float16 quantization. These results are summarized in Table \ref{tab:quantization}.

\begin{table*}[htbp]
\centering
\caption{Model Size and Performance Metrics with Different Quantization Techniques}
\begin{tabular}{|l|c|c|c|c|}
\hline
\textbf{Quantization} & \textbf{No Quantization} & \textbf{Float16 Quantization} & \textbf{Dynamic Range Quantization} & \textbf{UInt8 Quantization} \\ \hline
Size of Model (bytes) & 456768 & 235932 & 122848 & 124720 \\ \hline
Train Accuracy & 96\% & 96\% & 96\% & 17\% \\ \hline
Train Data Time (sample, inference) & 0.29 s & 0.29 s & 0.33 s & 0.27 s \\ \hline
Test Accuracy & 92\% & 92\% & 91\% & 17\% \\ \hline
Test Data Time (sample, inference) & 0.32 s & 0.31 s & 0.34 s & 0.28 s \\ \hline
\end{tabular}
\label{tab:quantization}
\end{table*}
\subsection{Performance Evaluation on Raspberry Pi}
To illustrate our approach, we utilized the tflite model deployed on the Raspberry Pi to evaluate training and testing accuracy as well as inference times. For the training data, we selected a total of 600 frames. The inference time averaged at 9.2 ms per sample. The accuracy percentage achieved was 86\%, slightly lower than the baseline performance observed over 20 epochs (refer to the confusion matrix in Figure \ref{fig:confusion}(c)). For the test set, the inference time remained consistent, while there was a marginal accuracy improvement of 0.1\%. In terms of power consumption, our desktop system (with GPU) has a wattage of 420W, compared to \textless5W for a Raspberry Pi \cite{b12}.

\subsection{Demonstrations}
There are two demonstrations which can be seen at this link: \url{https://youtu.be/tHTCzyHD7ak}. 

In the first demonstration, ultrasound data is downsampled on the Windows end before being transferred over UDP to the Raspberry Pi. To ensure reliable UDP data transfer, an artificial delay of 0.1 s was introduced. We achieved an accuracy of 85.3\% on the test set with an end-to-end inference time of 172 ms per frame.

In the second demonstration, we plotted 10 arbitrarily selected ultrasound images on both Windows and Raspberry Pi before and after the transfer. While this increased the latency significantly, it was essential to visually demonstrate the transfer of ultrasound data from Windows to Raspberry Pi. Despite the increased latency, the demonstration clearly shows the frames being transmitted. Of the 10 sampled transfers, 2 misclassifications resulted in an 80\% classification accuracy. On average, it took approximately 2 seconds per sample.

The significant reduction in model size achieved through quantization techniques is crucial for wearable applications, where memory and computational resources are limited. Our findings show that it is possible to maintain high accuracy and low latency, even with substantial model compression. Compared to previous studies that required powerful desktop GPUs, our approach demonstrates that real-time ultrasound-based gesture recognition is feasible on edge devices, making it a viable option for practical applications in healthcare and human-computer interaction.

\section{Limitations and Future Work}
This study marks the initial steps toward deploying a deep neural network on a Raspberry Pi for wearable ultrasound-based gesture recognition. While promising, the results indicate the start of broader research in this area. Key future directions include:

\begin{enumerate}
    \item \textbf{Platform Dependency:} Reliance on Windows-based hardware for ultrasound data acquisition is a limitation. Exploring alternative platforms like iOS or Android could improve wearability and accessibility.
    
    \item \textbf{Model Architecture Exploration:} This study did not explore transformer models or contemporary techniques for image classification and time-series analysis. Future work should investigate newer architectures and transfer learning to enhance classification performance.
\end{enumerate}

Moreover, future research could emphasize continual learning, subject-specific customization, and deploying models on smaller edge devices. These advancements are vital for expanding the usability of wearable ultrasound systems beyond gesture classification, into areas like health monitoring and remote diagnosis.

\section{Conclusions}
In conclusion, this study successfully demonstrates the deployment of a compressed deep learning model for ultrasound-based gesture recognition on a Raspberry Pi. The results highlight the potential for creating wearable, real-time gesture recognition systems with high accuracy and efficiency. Future work will focus on exploring more advanced model architectures, enhancing model personalization, and expanding the application scope to other domains such as remote health monitoring and assistive technologies.

\end{document}